\documentclass[lettersize,journal]{IEEEtran}
\usepackage{amsmath,amsfonts}
\usepackage{algorithmic}
\usepackage{algorithm}
\usepackage{array}
\usepackage[caption=false,font=normalsize,labelfont=sf,textfont=sf]{subfig}
\usepackage{textcomp}
\usepackage{stfloats}
\usepackage{url}
\usepackage{verbatim}
\usepackage{graphicx}
\usepackage{cite}

\usepackage{booktabs}
\usepackage{multirow}

\begin{document}

\title{A High-Accuracy Optical Music Recognition Method Based on Bottleneck Residual Convolutions}

\author{Junwen~Ma,
        Huhu~Xue,
        Xingyuan~Zhao,
        and~Weicheng~Fu%
\thanks{J. Ma and W. Fu are with the Department of Artificial Intelligence,
Tianshui Normal University, Tianshui 741001, Gansu, China.}%
\thanks{J. Ma and W. Fu are also with the Department of Physics,
Tianshui Normal University, Tianshui 741001, Gansu, China.}%
\thanks{H. Xue is with the Department of Artificial Intelligence,
Tianshui Normal University, Tianshui 741001, Gansu, China,
and also with the Department of Physics,
Tianshui Normal University, Tianshui 741001, Gansu, China.}%
\thanks{X. Zhao is with the School of Music and Dance,
Tianshui Normal University, Tianshui 741001, Gansu, China.}%
\thanks{W. Fu is the corresponding author (email: fuweicheng@tsnu.edu.cn).}%
}

\maketitle

\begin{abstract}

Optical Music Recognition (OMR) aims to convert printed or handwritten music score images into editable symbolic representations. This paper presents an end-to-end OMR framework that combines residual bottleneck convolutions with bidirectional gated recurrent unit (BiGRU)-based sequence modeling. A convolutional neural network with ResNet-v2-style residual bottleneck blocks and multi-scale dilated convolutions is used to extract features that encode both fine-grained symbol details and global staff-line structures. The extracted feature sequences are then fed into a BiGRU network to model temporal dependencies among musical symbols. The model is trained using the Connectionist Temporal Classification loss, enabling end-to-end prediction without explicit alignment annotations. Experimental results on the Camera-PrIMuS and PrIMuS datasets demonstrate the effectiveness of the proposed framework. On Camera-PrIMuS, the proposed method achieves a sequence error rate (SeER) of $7.52\%$ and a symbol error rate (SyER) of $0.45\%$, with pitch, type, and note accuracies of $99.33\%$, $99.60\%$, and $99.28\%$, respectively. The average training time is 1.74~s per epoch, demonstrating high computational efficiency while maintaining strong recognition performance. On PrIMuS, the method achieves a SeER of $8.11\%$ and a SyER of $0.49\%$, with pitch, type, and note accuracies of $99.27\%$, $99.58\%$, and $99.21\%$, respectively. A fine-grained error analysis further confirms the effectiveness of the proposed model.

\end{abstract}


\section{INTRODUCTION}\label{sec:level1}

Music, as an essential form of cultural heritage, has evolved continuously in both expressive form and dissemination medium throughout history. From medieval handwritten manuscripts to modern printed scores, musical notation has long served as a structured medium for preserving composers' artistic intent and musical semantics \cite{taruskin2006music}. In the digital era, the growing demand for large-scale preservation, retrieval, and analysis of musical content has significantly increased the importance of efficient music document digitization, making it a common concern across musicology, information science, and digital humanities \cite{Downie2003}. However, conventional music score archiving and analysis processes remain highly labor-intensive and time-consuming, which severely limits scalability and accessibility in digital music libraries \cite{Terras2010DigitalCR}.

Optical Music Recognition (OMR) aims to convert music score images into structured, computer-readable symbolic representations \cite{Bainbridge2001}, and plays a critical role in the automated preservation, analysis, and dissemination of musical information. Early OMR research, dating back to the mid-20th century, primarily focused on recognizing isolated and relatively simple musical symbols \cite{byrd2015towards}. With advances in computer vision and pattern recognition, OMR systems have gradually evolved from rule-based pipelines toward data-driven learning paradigms.

Traditional OMR systems typically rely on multi-stage processing pipelines \cite{Rebelo2012,byrd2015towards}, decomposing the recognition task into subtasks such as image preprocessing, staff-line detection, and symbol classification \cite{Pinto2011,Campos2016,vigliensoni2013optical}. Although these approaches achieved reasonable performance under controlled conditions, their dependence on handcrafted rules limits robustness and scalability, especially for handwritten or degraded scores \cite{Pacha2017}.

The rapid development of deep learning has significantly advanced OMR research. End-to-end learning frameworks enable models to learn hierarchical representations directly from raw score images, reducing reliance on manually designed preprocessing stages \cite{CalvoZaragoza2017EndtoEndOM,Calvo2018,Arnau2020}. More recent studies have explored advanced architectures, including Transformer-based image-to-sequence models and full-page recognition frameworks, to handle increasingly complex score layouts \cite{Vila2022,Rios-Vila2025}. Despite these advances, effectively modeling both fine-grained visual structures and temporal dependencies remains a central challenge in OMR.

Music scores encode complex hierarchical visual patterns, such as staff-line arrangements and symbol groupings, while simultaneously exhibiting strong sequential dependencies along the time axis. Existing deep learning approaches generally address these challenges from two complementary perspectives: spatial feature modeling and temporal sequence modeling. Designing an efficient framework that balances these two aspects is critical for achieving high recognition accuracy.

Motivated by the success of residual and bottleneck architectures in visual representation learning \cite{Kaiming2016,He2016}, as well as the effectiveness of bidirectional recurrent models for sequence modeling \cite{Graves2009,Schuster1997}, this paper proposes an end-to-end optical music recognition framework that integrates bottleneck residual convolutional feature extraction with BiGRU-based temporal sequence modeling. A bottleneck residual convolutional module with multi-scale dilated convolutions is employed to capture fine-grained symbol details and global staff-line structures, while a BiGRU network models contextual dependencies within feature sequences. The entire framework is trained end-to-end using the Connectionist Temporal Classification (CTC) loss \cite{Graves2006}, eliminating the need for explicit alignment annotations.

The main contributions of this work are summarized as follows:

(1) An end-to-end OMR framework is proposed that combines bottleneck residual convolutional feature extraction with bidirectional sequence modeling, and delivers strong performance on public benchmark datasets.

(2) Comprehensive comparisons with representative state-of-the-art OMR methods are conducted on public benchmarks. The experimental results show that the proposed approach achieves competitive overall performance while providing superior symbol-level recognition accuracy.

(3) The proposed method is further evaluated using the Optical Music Recognition Normalized Edit Distance (OMR-NED) metric \cite{martinezsevilla2025sheetmusicbenchmarkstandardized} on the PrIMuS and Camera-PrIMuS datasets, together with a fine-grained error analysis across different symbol categories.

\section{RELATED WORK}

\subsection{Traditional Optical Music Recognition}

Early optical music recognition systems predominantly relied on rule-based and multi-stage processing pipelines. These approaches decomposed the OMR task into a sequence of subtasks, including image preprocessing, staff-line detection and removal, and symbol segmentation and classification \cite{Rebelo2012,byrd2015towards,Pinto2011,Campos2016,vigliensoni2013optical}. While such methods achieved reasonable performance under controlled conditions, their heavy reliance on handcrafted rules and heuristic assumptions limited adaptability when handling diverse score layouts, degraded documents, or handwritten music notation \cite{Pacha2017}. In addition, error propagation across pipeline stages further constrained overall recognition performance.

\subsection{Deep Learning-Based End-to-End OMR}

The rapid development of deep learning techniques has significantly advanced optical music recognition research \cite{Calvo2023,castellanos2025deep}. End-to-end learning frameworks enable models to directly learn hierarchical representations from raw score images, substantially reducing reliance on manually designed preprocessing stages \cite{MAO2025,edirisooriya2021,vanderwel2017,Calvo_Zaragoza_2020,LUO2024,YU202084,ALFAROCONTRERAS2022,Calvo2018,CalvoZaragoza2017EndtoEndOM,liu2021,Arnau2020,Yipeng2023,Yu2024}.

These approaches have demonstrated notable improvements in symbol-level and sequence-level recognition accuracy. However, many existing methods rely on conventional convolutional backbones, which may struggle to simultaneously capture fine-grained musical symbol details and global structural patterns, such as staff-line configurations, under limited computational budgets.

\subsection{Advanced Architectures for OMR}

Recent studies have explored more advanced architectures to address complex score layouts and long-range dependencies in optical music recognition. Transformer-based image-to-sequence models and full-page recognition frameworks have been proposed to directly map score images to symbolic representations, enabling OMR systems to handle increasingly complex layouts and handwritten scores \cite{Vila2022,ríosvila2024,Rios-Vila2025,ríosvila2025}.

In addition, multimodal learning strategies and self-supervised or few-shot transfer methods have been investigated to improve generalization under limited annotated data scenarios \cite{Rosell2024,shatri2024,Penarrubia2024,Vila2023}. Despite their effectiveness, these approaches often involve increased computational complexity and data requirements, which may limit their applicability in resource-constrained settings.

\subsection{Summary and Motivation}

Although existing OMR methods have achieved notable progress, effectively balancing fine-grained visual representation, temporal dependency modeling, and computational efficiency remains challenging. Prior work often emphasizes either spatial feature extraction or sequence modeling, while their efficient integration has not been sufficiently explored.

Motivated by the strengths of bottleneck residual networks in visual representation learning and bidirectional recurrent models in sequence modeling, this work investigates an end-to-end OMR framework that integrates these components in a unified architecture. The proposed approach aims to achieve high recognition accuracy while maintaining an efficient and compact network design.

\section{PROPOSED METHOD}\label{sec2}
\subsection{Architecture}

\begin{figure*}[t]
    \centering
    \includegraphics[width=1.9\columnwidth]{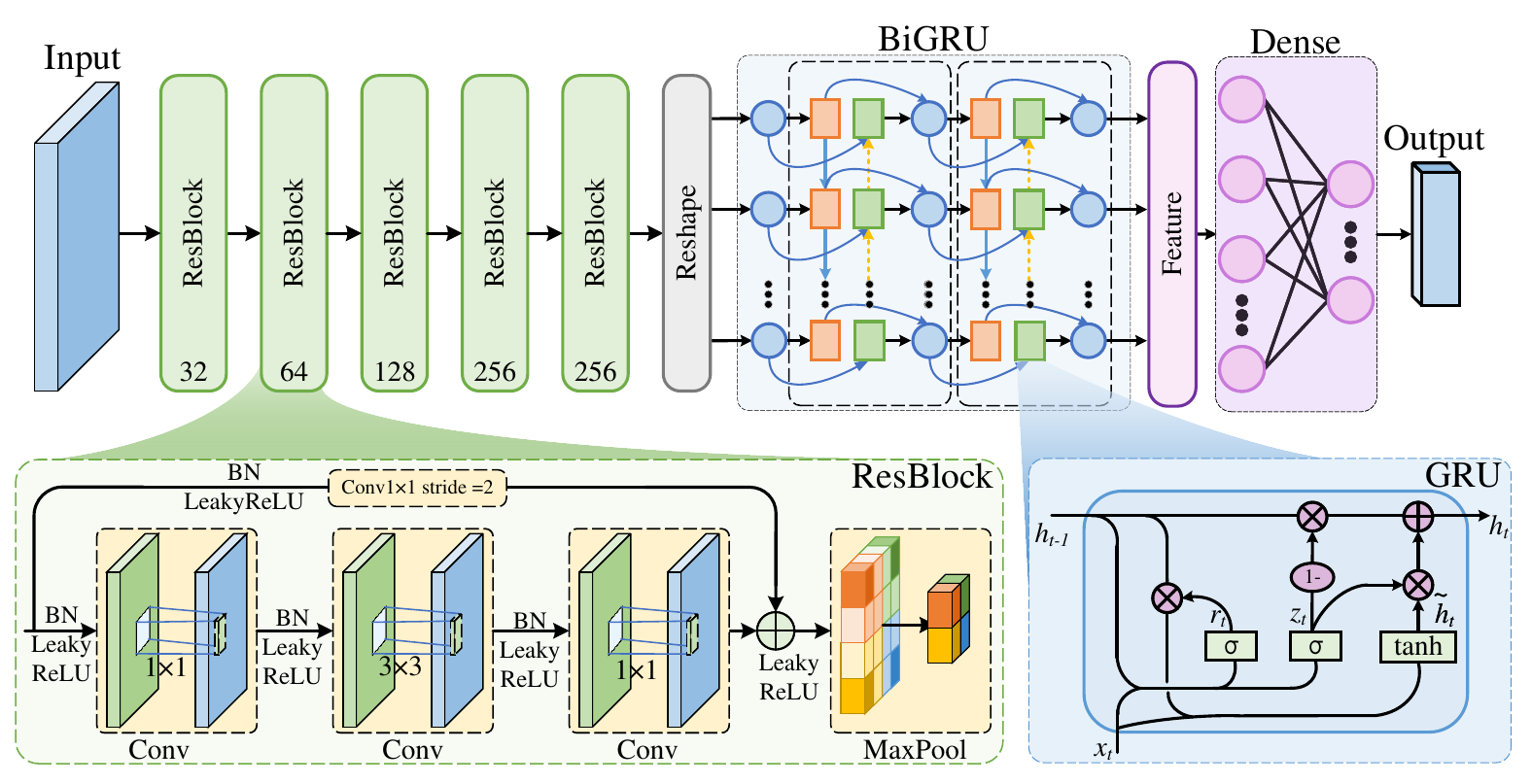}
    \caption{Architecture of the proposed end-to-end Optical Music Recognition model based on bottleneck residual convolution and BiGRU. The model consists of a bottleneck residual CNN feature extractor, a BiGRU sequence modeling module, and a fully connected output layer for end-to-end symbol sequence recognition.}\label{fig:model}
\end{figure*}

The proposed OMR model adopts an end-to-end architecture composed of three main components: a convolutional feature encoder, a BiGRU-based sequence modeling module, and a output layer. The overall design aims to jointly learn visual representations and sequential dependencies from music score images without requiring explicit symbol-level alignment. An overview of the proposed architecture is illustrated in Fig.~\ref{fig:model}.

Specifically, the convolutional feature encoder is built upon a deep bottleneck residual convolutional network, which is designed to extract both fine-grained local details, such as noteheads and accidentals, and global structural patterns, including staff-line configurations. The extracted two-dimensional feature maps are subsequently reshaped into a one-dimensional feature sequence, which serves as the input to the BiGRU module. The BiGRU network models contextual dependencies along the temporal dimension, capturing relationships between adjacent and long-range musical symbols. Finally, the output layer produces symbol probability sequences, which are optimized using the CTC criterion to enable alignment-free sequence prediction and end-to-end optical music recognition.

\subsection{Method}

To achieve end-to-end recognition, the proposed model comprises three key components: bottleneck residual convolutional feature extraction, Bi GRU-based sequence modeling, and a CTC output layer. The structure and underlying principles of each module are described in detail below.

\subsubsection{Bottleneck Residual Convolutional Feature Extraction}

The bottleneck residual convolutional module is employed to extract hierarchical spatial features from input music score images. This module follows the ResNet-v2 pre-activation bottleneck design \cite{He2016}, which consists of a sequence of $1 \times 1$ convolution for channel dimensionality reduction, a $3 \times 3$ dilated convolution for spatial feature extraction, and a $1 \times 1$ convolution for channel dimensionality restoration. The use of dilated convolutions enlarges the receptive field without increasing the number of parameters, enabling the model to capture both fine-grained symbol details and global staff-line structures.

The bottleneck design significantly reduces computational complexity while preserving representational capacity, making it suitable for efficient feature extraction in high-resolution music score images. Residual connections are incorporated to facilitate gradient propagation and stabilize the training of deep convolutional networks.

\subsubsection{BiGRU Sequence Modeling}

Following the convolutional feature extraction stage, the resulting two-dimensional feature maps are reshaped into a one-dimensional feature sequence and fed into a BiGRU module for sequence modeling. The BiGRU is employed to capture temporal dependencies along the sequence dimension by processing the feature sequence in both forward and backward directions. This bidirectional modeling allows the network to incorporate contextual information from past and future symbols, which is essential for representing temporal relationships in music scores.

At each time step, the GRU unit updates its hidden state by regulating information flow through update and reset gates, which control the contribution of the current input and the previous hidden state. This gating mechanism enables effective modeling of long-range dependencies while mitigating gradient degradation during training \cite{Schuster1997}. By combining forward and backward hidden representations, the BiGRU provides a context-aware feature representation for each time step in the sequence.

The output of the BiGRU module is subsequently passed through a fully connected layer followed by a softmax function to produce a probability distribution over symbol classes at each time step. These context-enhanced symbol probabilities serve as the input to the sequence prediction stage, facilitating accurate symbol recognition under the end-to-end learning framework.

\subsubsection{CTC Loss and Decoding}

To enable end-to-end training without explicit frame-level alignment, the proposed model is optimized using the CTC loss \cite{Graves2006}. The CTC objective allows the network to learn an implicit alignment between the input feature sequence and the target symbol sequence, making it well suited for optical music recognition, where symbol durations and spacing can vary significantly.

The CTC formulation augments the output symbol set with a special blank symbol, which enables flexible many-to-one and one-to-many mappings between input frames and output symbols. Given the symbol probability sequence produced by the output layer, the conditional probability of a target label sequence is computed by marginalizing over all valid alignment paths. During training, the model parameters are optimized by minimizing the negative log-likelihood of the target sequence.

During inference, the final symbol sequence is obtained through CTC decoding by collapsing repeated symbols and removing blank labels from the predicted frame-level outputs. This decoding process produces a symbol sequence without requiring explicit temporal alignment, completing the end-to-end optical music recognition pipeline.

\section{EXPERIMENTS}

\subsection{Datasets}
The proposed model is evaluated on two public benchmark datasets, PrIMuS and Camera-PrIMuS, which are widely used in optical music recognition research. These datasets cover a range of music score images from standardized printed notation to real-world photographs captured by cameras, enabling comprehensive evaluation under both controlled and challenging conditions.

The PrIMuS dataset provides music score images with a standardized layout and corresponding MusicXML annotations, making it well suited for supervised learning and quantitative evaluation of symbol recognition performance. The Camera-PrIMuS dataset extends PrIMuS by introducing real-world imaging artifacts, including noise, illumination variations, and perspective distortions, thereby more closely reflecting practical application scenarios.

\subsection{Implementation Details}
During training, the Adam optimizer is employed with an initial learning rate of $1 \times 10^{-4}$. A cosine annealing strategy is adopted to dynamically adjust the learning rate throughout training. The batch size is set to 16, and the maximum number of training iterations is 64{,}000.

The convolutional feature encoder consists of layers with channel dimensions of $[32, 64, 128, 256, 256]$. The hidden size of each BiGRU layer is set to 256. Additional training hyperparameters are summarized in Table~\ref{tab:hyperparams}.

\subsection{Evaluation Metrics}
To comprehensively assess the performance of the proposed model, six evaluation metrics are adopted:

\begin{enumerate}
  \item Sequence Error Rate (SeER): The ratio of incorrectly predicted sequences to the total number of sequences. A sequence is considered incorrect if it contains at least one error in notes, pitches, rests, or other musical symbols.

  \item Symbol Error Rate (SyER): The average number of editing operations, including insertions, substitutions, and deletions, required to transform the predicted sequence into the ground-truth sequence, normalized by the sequence length.

  \item Pitch Accuracy: The proportion of notes for which the pitch is correctly predicted.

  \item Note Type Accuracy: The proportion of notes for which the note type is correctly predicted.

  \item Note Accuracy: The proportion of notes for which both pitch and note type are correctly predicted.

  \item OMR Normalized Edit Distance (OMR-NED):
  Proposed by Martinez-Sevilla \textit{et al.}~\cite{martinezsevilla2025sheetmusicbenchmarkstandardized}, OMR-NED evaluates recognition quality from the perspective of visual music symbols. It measures the normalized number of insertions and deletions between the predicted score and the ground-truth score:
  \begin{equation}\label{eq-OMR_NED}
    \text{OMR-NED} = \frac{I + D}{N_1 + N_2},
  \end{equation}
  where $I$ and $D$ denote the numbers of inserted and deleted symbols, respectively, and $N_1$ and $N_2$ denote the total number of symbols in the predicted score and the ground-truth score. Compared with traditional SER-based metrics, OMR-NED enables a more fine-grained analysis of symbol-level errors, such as pitch, rhythmic, and structural deviations.
\end{enumerate}

These metrics provide a multi-level evaluation of OMR performance. SeER and SyER assess recognition quality at the sequence level, while pitch accuracy, note type accuracy, and note accuracy reflect symbol-level recognition performance. In addition, OMR-NED offers further insight into fine-grained recognition errors by explicitly quantifying deviations in musical symbol structure, enabling more interpretable evaluation of OMR system behavior.

\subsection{Image Preprocessing}

\begin{table}[t]
\centering
\caption{Data augmentation operations and activation probabilities}
\label{tab:augmentation}
\renewcommand{\arraystretch}{1.25}
\begin{tabular}{@{}ll@{}}
\toprule
\textbf{Augmentation Type} & \textbf{Probability} \\
\midrule
Brightness/Contrast/Gamma Adjustment & 0.9 \\
Blur Perturbation & 0.5 \\
JPEG Compression Artifacts & 0.5 \\
Morphological Erosion & 0.08 \\
Morphological Dilation & 0.08 \\
Small-Angle Rotation and Shearing & 0.6 \\
Perlin Noise & 0.25 \\
Gaussian Noise & 0.25 \\
Elastic Deformation & 0.15 \\
Scratches & 0.05 \\
\bottomrule
\end{tabular}
\end{table}

\begin{figure}[htbp]
    \centering
    \includegraphics[width=.92\columnwidth]{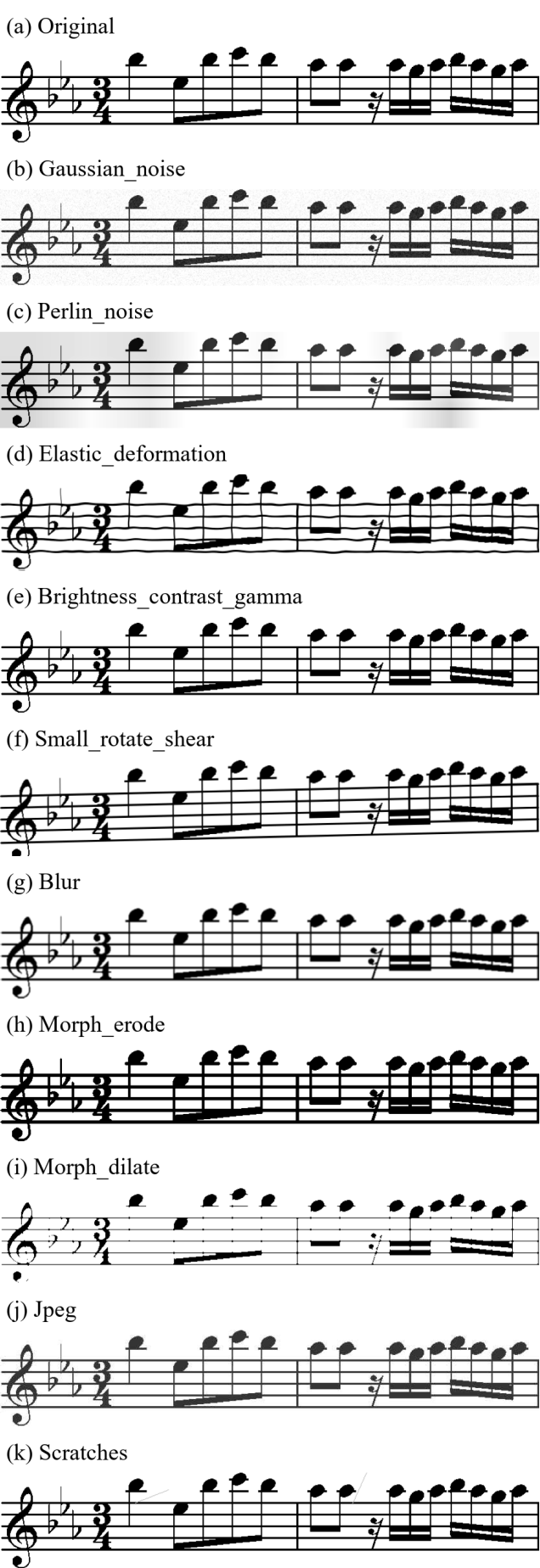}
    \caption{Comparison between the original music score image and its augmented versions.}
    \label{fig:2}
\end{figure}

Prior to training, all input sheet music images are resized to a fixed height of $H = 128$, while the image width is scaled proportionally to preserve the original aspect ratio. This normalization ensures consistent input resolution while maintaining the spatial structure of musical symbols and staff lines.

To improve generalization under diverse imaging conditions, a lightweight data augmentation strategy is applied during training. The augmentation module introduces random perturbations that simulate common degradations encountered in real-world music score acquisition, including brightness and contrast variations, local geometric distortions, noise injection, and compression artifacts. These transformations are designed to preserve the structural integrity of staff lines and musical symbols while increasing variability in visual appearance. Representative augmentation examples are illustrated in Fig.~\ref{fig:2}.

Each augmentation operation is activated independently with a predefined probability. When multiple activation conditions are satisfied, an input image may undergo multiple perturbations within a single training iteration, resulting in composite degraded samples. This strategy simulates the concurrent degradation patterns commonly observed in scanned or camera-captured sheet music images, enabling the model to better adapt to complex image variations without altering the underlying musical content.

\begin{table*}[t]
\centering
\caption{Main hyperparameter configuration of the model.}
\label{tab:hyperparams}
\begin{tabular}{lcc}
\toprule
\textbf{Category} & \textbf{Name} & \textbf{Value} \\
\midrule
Input & Image height & 128 \\
 & Number of channels & 1 \\
 & Batch size & 16 \\
\midrule
Convolutional encoder & Number of residual blocks & 5 \\
 & Convolution channels & [32, 64, 128, 256, 256] \\
 & Kernel size & $3\times3$ \\
 & Dilation rate & $(1,1),(2,1),(4,1),(8,1),(1,1)$ \\
 & Pooling window & $[2\times2, 2\times2, 2\times1, 2\times1, 1\times1]$ \\
 & Bottleneck ratio & [4, 4, 4, 4, 4] \\
\midrule
BiGRU & Number of layers & 2 \\
 & Units per layer & 256 \\
\bottomrule
\end{tabular}
\end{table*}

\begin{table*}[htbp]
\centering
\caption{Main hyperparameter configuration of the model}
\label{tab:hyperparams}
\begin{tabular}{lcc}
\toprule
\textbf{Category} & \textbf{Name} & \textbf{Value} \\
\midrule
Input
    & Img.\ height & 128 \\
    & Num.\ channels & 1 \\
    & Batch size & 16 \\
\midrule
Conv.\ encoder
    & Num.\ residual blocks & 5 \\
    & Conv.\ channels & [32, 64, 128, 256, 256] \\
    & Kernel size & $3\times3$ \\
    & Dilation rate & $(1,1),(2,1),(4,1),(8,1),(1,1)$ \\
    & Pool window & $[2\times2, 2\times2, 2\times1, 2\times1, 1\times1]$ \\
    & Bottleneck ratio & [4, 4, 4, 4, 4] \\
\midrule
BiGRU
    & Num.\ layers & 2 \\
    & Hidden size & 256 \\
\bottomrule
\end{tabular}
\end{table*}

\subsection{Experimental Results and Analysis}

\subsubsection{Comparison with Baseline Models}

To evaluate the effectiveness of the proposed architecture, comparative experiments are conducted on the Camera-PrIMuS and PrIMuS datasets against two representative baseline models, namely CRNN~\cite{CalvoZaragoza2018CameraPrIMuSNE} and R2-CRNN~\cite{liu2021}. To ensure a fair comparison, no data augmentation strategies are applied to any of the evaluated models. The quantitative results under different training and validation settings are reported in Tables~\ref{tab:3} and~\ref{tab:primus_baseline}, where each result is presented in the form of ``SyER / SeER'', and lower values indicate better recognition performance.

As shown in Table~\ref{tab:3}, the proposed Bottleneck ResNet-BiGRU model demonstrates stable and competitive performance under different training--validation configurations. When trained on the Clean subset and evaluated under the Clean condition, the model achieves SyER/SeER values of 0.49\% / 8.66\% and 0.49\% / 7.80\% under the Agnostic and Semantic encodings, respectively. Compared with CRNN and R2-CRNN, the proposed method yields lower error rates in this setting. When trained and evaluated on the Distortions subset, the Bottleneck ResNet-BiGRU model also achieves the lowest SyER and SeER among all compared methods. In the cross-condition setting, where the model is trained on Clean data and evaluated on Distortions data, the proposed method shows slightly higher error rates than R2-CRNN, while still outperforming CRNN in both SyER and SeER. Overall, these results indicate that the proposed model is able to maintain stable performance across different data conditions on the Camera-PrIMuS benchmark.

Table~\ref{tab:primus_baseline} presents the comparison of recognition performance between the proposed model and the CRNN baseline on the PrIMuS dataset. The experimental results show that the proposed method significantly outperforms the CRNN baseline under both encoding schemes.

Specifically, under the Agnostic encoding setting, the Symbol Error Rate (SyER) of the proposed model is reduced from 1.00\% for CRNN to 0.53\%, while the Sequence Error Rate (SeER) decreases from 17.9\% to 9.09\%. Under the Semantic encoding setting, the proposed model also achieves superior recognition performance, with SyER reduced from 0.80\% to 0.50\% and SeER reduced from 12.5\% to 8.33\%. These results demonstrate that the proposed method consistently achieves better recognition performance than the baseline model under both encoding schemes.

\begin{table*}[t]
\centering
\caption{Recognition performance of different models on the Camera-PrIMuS dataset under Clean and Distortions conditions (without data augmentation, in \%)}
\label{tab:3}
\begin{tabular}{llcccc}
\toprule
\textbf{Data type} & \textbf{Model} & \multicolumn{2}{c}{\textbf{Clean}} & \multicolumn{2}{c}{\textbf{Distortions}} \\
\cmidrule(lr){3-4} \cmidrule(lr){5-6}
 &  & Agnostic & Semantic & Agnostic & Semantic \\
\midrule
\multirow{3}{*}{Clean}
 & CRNN & 1.10 / 21.7 & 0.80 / 12.5 & 44.3 / 95.1 & 59.7 / 97.9 \\
 & R2-CRNN & 0.59 / 16.2 & 0.56 / 13.4 & 7.63 / 20.9 & 38.19 / 91.0 \\
 & Ours & 0.49 / 8.66 & 0.49 / 7.80 & 23.41 / 87.91 & 39.05 / 93.6 \\
\midrule
\multirow{3}{*}{Distortions}
 & CRNN & 1.40 / 24.9 & 3.30 / 44.6 & 1.60 / 24.7 & 3.40 / 38.3 \\
 & R2-CRNN & 0.68 / 18.7 & 1.06 / 25.4 & 1.01 / 27.8 & 1.50 / 35.8 \\
 & Ours & 0.63 / 10.73 & 0.79 / 11.06 & 0.83 / 13.9 & 1.27 / 16.94 \\
\bottomrule
\end{tabular}
\end{table*}

\begin{table}[htbp]
\centering
\caption{Recognition performance of different models on the PrIMuS dataset under Clean and Distortions conditions (without data augmentation, in \%)}
\label{tab:primus_baseline}
\renewcommand{\arraystretch}{1.2}
\begin{tabular}{lcc}
\toprule
\textbf{Model} & \textbf{Agnostic} & \textbf{Semantic} \\
\midrule
CRNN & 1.00 / 17.9 & 0.80 / 12.5 \\
Ours & 0.53 / 9.09 & 0.50 / 8.33 \\
\bottomrule
\end{tabular}
\end{table}

\subsubsection{Effectiveness of the Data Augmentation Strategy}

To assess the impact of the proposed data augmentation strategy, we compare the performance of the model trained with and without data augmentation on both the PrIMuS and Camera-PrIMuS datasets. The quantitative results are reported in Tables~\ref{tab:6} and \ref{tab:7}. Overall, data augmentation yields performance improvements on both datasets, although the magnitude of improvement varies across different encoding schemes and data characteristics.

On the Camera-PrIMuS dataset, data augmentation consistently improves recognition performance under both encoding schemes. For the Semantic encoding, the introduction of augmentation reduces the SeER from $7.80\%$ to $7.52\%$ and the SyER from $0.49\%$ to $0.45\%$. Under the Agnostic encoding, the improvement is more pronounced, with the SeER decreasing from $8.66\%$ to $7.64\%$ and the SyER from $0.49\%$ to $0.44\%$. These results indicate that the augmentation strategy effectively enhances model generalization in the presence of real-world image degradations.

\begin{table}[t]
\centering
\caption{Performance comparison of our model with and without data augmentation on the Camera-PrIMuS dataset.}
\renewcommand{\arraystretch}{1.2}
\begin{tabular}{lcc}
\toprule
\textbf{Methods} & \textbf{SeER (\%)} & \textbf{SyER (\%)} \\
\midrule
Semantic & 7.80 & 0.49 \\
Semantic+Augmentation & 7.52 & 0.45 \\
Agnostic & 8.66 & 0.49 \\
Agnostic+Augmentation & 7.64 & 0.44 \\
\bottomrule
\end{tabular}
\label{tab:6}
\end{table}

\begin{table}[t]
\centering
\caption{Performance comparison of our model with and without data augmentation on the PrIMuS dataset}
\renewcommand{\arraystretch}{1.2}
\begin{tabular}{lcc}
\toprule
\textbf{Methods} & \textbf{SeER (\%)} & \textbf{SyER (\%)} \\
\midrule
Semantic & 8.33 & 0.50 \\
Semantic+Augmentation & 8.11 & 0.49 \\
Agnostic & 9.09 & 0.53 \\
Agnostic+Augmentation & 8.55 & 0.48 \\
\bottomrule
\end{tabular}
\label{tab:7}
\end{table}

In contrast, the impact of data augmentation on the PrIMuS dataset exhibits a stronger dependence on the encoding scheme. For the Semantic encoding, augmentation leads to a modest reduction in SeER from $8.33\%$ to $8.11\%$ and in SyER from $0.50\%$ to $0.49\%$. This limited improvement suggests that the clean and structurally standardized nature of PrIMuS leaves relatively little room for further enhancement through visual perturbations. Conversely, under the Agnostic encoding, data augmentation results in more noticeable gains, with the SeER decreasing from $9.09\%$ to $8.55\%$ and the SyER from $0.53\%$ to $0.48\%$, indicating that augmentation helps mitigate symbol-level ambiguity in less constrained representations.

Overall, data augmentation primarily contributes to reductions in sequence-level error rates, with the most consistent and substantial improvements observed on the Camera-PrIMuS dataset, which contains more complex and realistic imaging conditions.

\subsubsection{Comparison with State-of-the-Art (SOTA) Methods}

To further analyze the recognition capability of the proposed model, we compute SER, SyER, Pitch Accuracy, and Duration Accuracy under the semantic encoding on both the PrIMuS and Camera-PrIMuS datasets, and compare the results with three representative state-of-the-art methods: MCRR \cite{Yu2024}, ConvNet-STN \cite{Choi2017BootstrappingSO}, and DWD \cite{Zhang2023ADF}. The comparison results are reported in Tables~\ref{tab:4} and \ref{tab:5}.

\begin{table*}[t]
\centering
\caption{ Performance comparison of SOTA models on the Camera-PrIMuS dataset}
\renewcommand{\arraystretch}{1.2}
\resizebox{\textwidth}{!}{%
\begin{tabular}{lcccccc}
\toprule
\textbf{Methods} & \textbf{SeER} & \textbf{SyER} & \textbf{Pitch accuracy} & \textbf{Type accuracy} & \textbf{Note accuracy} & \textbf{Training time (s/epoch)} \\
\midrule
ConvNet-STN & 27.0298 & 9.3975 & 70.0000 & 72.3987 & 68.3274 & 2.79 \\
DWD & 25.1475 & 7.9030 & 82.1123 & 80.9635 & 78.5064 & 3.50 \\
MCRR & 5.1488 & 1.0612 & 90.7544 & 91.5520 & 88.2094 & 1.93 \\
\textbf{Ours} & 7.52 & 0.45 & 99.33 & 99.60 & 99.28 & 1.74 \\
\bottomrule
\end{tabular}
} 
\label{tab:4}
\end{table*}

\begin{table*}[t]
\centering
\caption{Performance comparison of SOTA models on the PrIMuS dataset}
\renewcommand{\arraystretch}{1.2}
\resizebox{\textwidth}{!}{%
\begin{tabular}{lcccccc}
\toprule
\textbf{Methods} & \textbf{SeER} & \textbf{SyER} & \textbf{Pitch accuracy} & \textbf{Type accuracy} & \textbf{Note accuracy} & \textbf{Training time (s/epoch)} \\
\midrule
ConvNet-STN & 16.8056 & 5.0208 & 89.7107 & 94.0273 & 85.5620 & 0.98 \\
DWD & 18.5609 & 8.7811 & 95.2074 & 96.0049 & 91.1255 & 1.21 \\
MCRR & 1.4571 & 0.3234 & 97.0000 & 97.1298 & 94.4447 & 0.56 \\
\textbf{Ours} & 8.11 & 0.49 & 99.27 & 99.58 & 99.21 & 1.80 \\
\bottomrule
\end{tabular}
}
\label{tab:5}
\end{table*}

From the results in Table~\ref{tab:4}, it can be seen that, on the Camera-PrIMuS dataset, which better reflects real-world image capture conditions, the proposed model demonstrates particularly strong performance at the semantic level. Its Pitch Accuracy, Type Accuracy, and Note Accuracy reach 99.33\%, 99.60\%, and 99.28\%, respectively, all of which are higher than those of the other compared methods. In terms of symbol-level evaluation metrics, the proposed model achieves a SyER of 0.45\%, which is the best result among all methods, indicating that the model has high accuracy in overall symbol sequence matching. The SeER is 7.52\%, which, although slightly higher than that of MCRR, is still clearly better than those of ConvNet-STN and DWD. In addition, Table~\ref{tab:4} also shows that the average training time of the proposed model on Camera-PrIMuS is 1.74~s/epoch, which is lower than that of ConvNet-STN, DWD, and MCRR. In other words, while improving recognition performance, the model does not introduce excessive computational cost, thus maintaining a relatively reasonable balance between accuracy and efficiency.

From the results shown in Table~\ref{tab:5}, the proposed model achieves the best performance on the three core recognition metrics---Pitch accuracy, Type accuracy, and Note accuracy---with all accuracies exceeding $99\%$, significantly outperforming the other comparative methods. In terms of overall recognition quality, although the symbol-level error rates (SeER and SyER) are slightly higher than those of MCRR, they remain clearly superior to ConvNet-STN and DWD, demonstrating strong overall competitiveness.

It is worth noting that the average training time of the proposed model is 1.80 s/epoch, which is slightly slower than the other methods. However, this additional training cost yields substantial gains in recognition precision, as the proposed model achieves state-of-the-art performance across pitch, duration, and note-level classification tasks. Therefore, despite the marginally increased computational overhead, the stability and high accuracy of the model on these key tasks make it highly advantageous on this dataset.

\subsubsection{Fine-Grained Error Analysis with OMR-NED}

To further evaluate the model's recognition capability at the symbol level, this paper conducts a fine-grained error analysis on the Camera-PrIMuS and PrIMuS datasets using the OMR-NED metric. OMR-NED measures structural differences through insertion, deletion, and substitution operations, and compared with the traditional SER, it provides a more precise reflection of the difficulty associated with different symbol categories during recognition.

On the Camera\text{-}PrIMuS dataset (Tables~\ref{tab:8} and \ref{tab:9}), both encoding schemes achieve low overall OMR-NED scores, with $0.32\%$ under Semantic encoding and $0.28\%$ under Agnostic encoding. In terms of category distribution, Notes constitute the primary source of errors (Semantic: $35.19\%$, Agnostic: $15.90\%$), largely due to their overwhelming prevalence among all symbols. Meanwhile, GraceNotes exhibit notably higher OMR-NED values under both encodings (Semantic: $8.73\%$, Agnostic: $7.54\%$), making them one of the most challenging categories for the model. In addition, \textbf{Ties} (Semantic: $4.26\%$, Agnostic: $3.95\%$) and Accidentals (Agnostic: $0.72\%$, contributing $18.53\%$ of total errors) also show elevated error levels, indicating that these elongated or context-dependent symbols remain difficult to recognize accurately.

\begin{table*}[htbp]
\centering
\caption{Fine-grained OMR-NED analysis across different symbol categories (Semantic encoding) on the Camera-PrIMuS dataset}
\renewcommand{\arraystretch}{1.2}
\begin{tabular}{lrrrrrr}
\toprule
\textbf{Category} & \textbf{I} & \textbf{D} & \textbf{N1} & \textbf{N2} & \textbf{OMR-NED (\%)} & \textbf{ Errors (\%)} \\
\midrule
Barlines        & 76  & 190 & 29,516  & 29,402  & 0.45 & 22.77 \\
Clefs           & 3   & 3   & 7,935   & 7,935   & 0.04 & 0.51 \\
GraceNotes      & 186 & 172 & 2,043   & 2,057   & 8.73 & 30.65 \\
KeySignatures   & 2   & 0   & 6,095   & 6,097   & 0.02 & 0.17 \\
MultiRests      & 4   & 5   & 1,990   & 1,989   & 0.23 & 0.77 \\
Notes           & 202 & 209 & 120,423 & 120,416 & 0.17 & 35.19 \\
Rests           & 5   & 5   & 11,323  & 11,323  & 0.04 & 0.86 \\
Ties            & 35  & 62  & 1,152   & 1,125   & 4.26 & 8.30 \\
TimeSignatures  & 5   & 4   & 7,890   & 7,891   & 0.06 & 0.77 \\
\midrule
\textbf{Overall} & -- & -- & -- & -- & \textbf{0.32} & \textbf{SyER = 0.45\%} \\
\bottomrule
\end{tabular}
\label{tab:8}
\end{table*}

\begin{table*}[htbp]
\centering
\caption{Fine-grained OMR-NED analysis across different symbol categories (Agnostic encoding) on the Camera-PrIMuS dataset.}
\renewcommand{\arraystretch}{1.2}
\begin{tabular}{lrrrrrr}
\toprule
\textbf{Category} & \textbf{I} & \textbf{D} & \textbf{N1} & \textbf{N2} & \textbf{OMR-NED (\%)} & \textbf{ Errors (\%)} \\
\midrule
Accidentals      & 148 & 77  & 15,663  & 15,734  & 0.72 & 18.53 \\
Barlines         & 66  & 208 & 29,619  & 29,477  & 0.46 & 22.57 \\
Clefs            & 1   & 1   & 7,957   & 7,957   & 0.01 & 0.16 \\
Fermatas         & 1   & 0   & 394     & 395     & 0.13 & 0.08 \\
GraceNotes       & 154 & 152 & 2,028   & 2,030   & 7.54 & 25.21 \\
MeterSigns       & 0   & 1   & 4,230   & 4,229   & 0.01 & 0.08 \\
Notes            & 96  & 97  & 120,800 & 120,799 & 0.08 & 15.90 \\
Others           & 7   & 3   & 22,968  & 22,972  & 0.02 & 0.82 \\
Rests            & 2   & 2   & 11,465  & 11,465  & 0.02 & 0.33 \\
Slurs            & 89  & 109 & 2,518   & 2,498   & 3.95 & 16.31 \\
\midrule
\textbf{Overall} & -- & -- & -- & -- & \textbf{0.28} & \textbf{SyER = 0.46\%} \\
\bottomrule
\end{tabular}
\label{tab:9}
\end{table*}

\begin{table*}[htbp]
\centering
\caption{Fine-grained OMR-NED analysis across different symbol categories (Semantic encoding) on the PrIMuS dataset}
\renewcommand{\arraystretch}{1.2}
\begin{tabular}{lrrrrrr}
\toprule
\textbf{Category} & \textbf{I} & \textbf{D} & \textbf{N1} & \textbf{N2} & \textbf{OMR-NED (\%)} & \textbf{ Errors (\%)} \\
\midrule
Barlines        & 90  & 219 & 29,413  & 29,284  & 0.53 & 22.91 \\
Clefs           & 4   & 2   & 7,950   & 7,952   & 0.04 & 0.44 \\
GraceNotes      & 207 & 200 & 2,102   & 2,109   & 9.67 & 30.17 \\
KeySignatures   & 4   & 3   & 6,054   & 6,055   & 0.06 & 0.52 \\
MultiRests      & 5   & 5   & 1,947   & 1,947   & 0.26 & 0.74 \\
Notes           & 239 & 243 & 120,676 & 120,672 & 0.20 & 35.73 \\
Rests           & 4   & 2   & 11,410  & 11,412  & 0.03 & 0.44 \\
Ties            & 37  & 68  & 1,184   & 1,153   & 4.49 & 7.78 \\
TimeSignatures  & 9   & 8   & 7,892   & 7,893   & 0.11 & 1.26 \\
\midrule
\textbf{Overall} & -- & -- & -- & -- & \textbf{0.36} & \textbf{SyER = 0.49\%} \\
\bottomrule
\end{tabular}
\label{tab:10}
\end{table*}

\begin{table*}[htbp]
\centering
\caption{Fine-grained OMR-NED analysis across different symbol categories (Agnostic encoding) on the PrIMuS dataset}
\renewcommand{\arraystretch}{1.2}

\begin{tabular}{lrrrrrr}
\toprule
\textbf{Category} & \textbf{I} & \textbf{D} & \textbf{N1} & \textbf{N2} & \textbf{OMR-NED (\%)} & \textbf{ Errors (\%)} \\
\midrule
Accidentals      & 140 & 61  & 15,415  & 15,494  & 0.65 & 16.24 \\
Barlines         & 93  & 178 & 29,347  & 29,262  & 0.46 & 21.89 \\
Clefs            & 1   & 0   & 7,953   & 7,954   & 0.01 & 0.08 \\
Fermatas         & 5   & 0   & 406     & 411     & 0.61 & 0.40 \\
GraceNotes       & 210 & 201 & 2,232   & 2,241   & 9.19 & 33.20 \\
MeterSigns       & 0   & 0   & 4,233   & 4,233   & 0.00 & 0.00 \\
Notes            & 65  & 70  & 120,637 & 120,632 & 0.06 & 10.90 \\
Others           & 13  & 8   & 22,477  & 22,482  & 0.05 & 1.70 \\
Rests            & 0   & 0   & 11,446  & 11,446  & 0.00 & 0.00 \\
Slurs            & 77  & 116 & 2,221   & 2,182   & 4.38 & 15.59 \\
\midrule
\textbf{Overall} & -- & -- & -- & -- & \textbf{0.29} & \textbf{SyER = 0.48\%} \\
\bottomrule
\end{tabular}
\label{tab:11}
\end{table*}

On the PrIMuS dataset (Tables~\ref{tab:10} and \ref{tab:11}), the overall OMR-NED is slightly higher than on Camera\text{-}PrIMuS but still remains low (Semantic: $0.36\%$, Agnostic: $0.29\%$). Although the images in this dataset are cleaner and free from camera distortions, the category-level error distribution exhibits similar tendencies: Notes continue to be the major source of errors (Semantic: $35.73\%$, Agnostic: $10.90\%$), while GraceNotes again show prominently high OMR-NED values (Semantic: $9.67\%$, Agnostic: $9.19\%$), highlighting the structural complexity of grace notes as a key factor affecting recognition accuracy. In contrast, fundamental structural symbols such as Clefs, Rests, and KeySignatures consistently maintain near-zero error rates on both datasets, demonstrating the model’s strong stability in recognizing common and structurally well-defined symbols.

Overall, the model achieves high accuracy in recognizing common structural symbols (e.g., notes, clefs, and time signatures), but it still faces challenges with categories that exhibit complex structures, occur infrequently, or rely heavily on contextual cues (such as GraceNotes, Accidentals, and Ties). Future improvements may focus on increasing the diversity of detailed-symbol samples, incorporating stronger local and structural context modeling mechanisms, or applying class-balanced strategies, with the aim of further enhancing the model's robustness at the fine-grained symbol level.

\section{CONCLUSION}

This paper presented an end-to-end optical music recognition framework that combines bottleneck residual convolutional feature extraction with bidirectional GRU-based sequence modeling. The proposed method was evaluated on monophonic music score benchmarks, including PrIMuS and Camera-PrIMuS, which cover both standardized printed scores and camera-captured images with real-world degradations.

Extensive experiments and multi-level evaluations show that the proposed method achieves competitive performance in both symbol-level and sequence-level recognition under semantic and agnostic encoding schemes. In particular, the method attains low symbol error rates and low OMR-NED, while achieving pitch- and note-level recognition accuracies above $99\%$ under semantic encoding. These results demonstrate the effectiveness of the proposed framework for optical music recognition.

The experimental results further show that the adopted data augmentation strategy improves generalization under challenging imaging conditions, particularly on the Camera-PrIMuS dataset. In addition, the fine-grained OMR-NED-based error analysis indicates that most of the remaining recognition errors are associated with structurally complex symbols, such as grace notes, suggesting potential directions for improving detailed symbol modeling.


\section*{Acknowledgments}

This work was supported by the National Natural Science Foundation of China (Grants Nos. 12465010, 12247106). W. Fu also acknowledges support from the Long-yuan Youth Talents Project of Gansu Province, the Fei-tian Scholars Project of Gansu Province, the Leading Talent Project of Tianshui City, and the Project of Open Competition for the Best Candidates from Department of Education of Gansu Province (Grant No. 2021jyjbgs-06).


\vfill

\bibliographystyle{elsarticle-num}

\end{document}